\documentclass[a4paper,11pt,twocolumn]{article}

\usepackage{icphs2023}
\usepackage{metalogo} 
\usepackage{epstopdf}
\usepackage{url}
\usepackage{multirow}

\title{What does BERT learn about prosody?}
\author{Sofoklis Kakouros\textsuperscript{a} and Johannah O'Mahony\textsuperscript{b}}
\organization{\textsuperscript{a}University of Helsinki, \textsuperscript{b}University of Edinburgh}
\email{sofoklis.kakouros@helsinki.fi, johannah.o'mahony@ed.ac.uk}
\begin{document}

\maketitle

\begin{abstract}
\vspace{-2mm}
Language models have become nearly ubiquitous in natural language processing applications achieving state-of-the-art results in many tasks including prosody. As the model design does not define predetermined linguistic targets during training but rather aims at learning generalized representations of the language, analyzing and interpreting the representations that models implicitly capture is important in bridging the gap between interpretability and model performance. Several studies have explored the linguistic information that models capture providing some insights on their representational capacity. However, the current studies have not explored whether prosody is part of the structural information of the language that models learn. In this work, we perform a series of experiments on BERT probing the representations captured at different layers. Our results show that information about prosodic prominence spans across many layers but is mostly focused in middle layers suggesting that BERT relies mostly on syntactic and semantic information.
\end{abstract}


\section{Introduction}
\vspace{-2mm}
Pre-trained language models (LMs) have achieved high performance on several natural language processing (NLP) tasks such as constituency parsing \cite{kitaev2018constituency}, semantic role labeling, and coreference resolution \cite{peters-etal-2018-deep}. Recently, prosody and prosodic phenomena have witnessed increasing attention in natural language applications enabled by pre-trained LMs leading to improved performance \cite{talman_etal2019prosody,stephenson2022bert}. Despite the gains, we do not yet fully understand what enables these models to perform at this level. Their performance suggests that their learning objectives potentially teach the models details about the structure of the language. It remains, however, unclear, what specific prosodic information these models are able to implicitly capture during their pre-training. In this work, we focus on analyzing the prosodic information in one of the most widespread models, BERT (Bidirectional Encoder Representations from Transformers) \cite{devlin2018bert}.

Several recent works have investigated the representations learned at different layers of BERT in an attempt to interpret and understand the linguistic information captured by the model. These studies have uncovered that BERT indeed learns various aspects of the language. For instance, \cite{jawahar2019does} showed that BERT layers capture a rich hierarchy of linguistic information spanning from surface features in the lower layers to syntactic in middle layers and semantic in the higher layers. These findings are further supported by a number of other works with the general observation indicating that learned representations vary with increasing network depth, with greater depth typically involving linguistic functions that require larger contextual relationships across the word tokens \cite{peters-etal-2018-deep,peters2018dissecting,tenney2019bert}. However, to the best of our knowledge, BERT has not been examined with respect to its prosodic information.

In general, prosody can be viewed as the characteristics in an utterance that extend individual phonetic segments and encapsulate phonetic and phonological properties that are not due to the choice of individual lexical items, but depend on factors such as their semantic and syntactic relations \cite{wagner2010experimental,kakouros2016perception}. These characteristics convey information about the meaning and structure of an utterance. Although prosody is a characteristic of spoken language, the prosodic patterns in speech are connected and interact with their associated sequences of syllables, words, and phrases. Therefore, it is meaningful to assume that some aspects of the prosodic variation can be captured from text alone \cite{kakouros2016analyzing, talman_etal2019prosody}. 

In this work, we investigate how prosodic information is linguistically encoded by probing BERT with respect to the prosodic phenomenon of prominence. Prosodic prominence is defined as the subjective impression of a linguistic unit standing out of its context \cite{kakouros20163pro,terken2000perception,kakouros2015automatic}. Given the recent success of BERT in predicting prosodic prominence \cite{talman_etal2019prosody}, we attempt to answer the question of what BERT learns about prosody during its pre-training. Does the model rely on general linguistic and syntactic knowledge for its prosodic predictions or does it have a different pattern of weight allocations across its layers suggesting that the model can capture prosodic information?

We use three datasets with prominence annotations and examine the weights at different layers and compare them with  existing findings from the literature on other tasks. To further validate our approach with the literature  we also extract part-of-speech tags for our data and examine the learned layer weights. The code to reproduce the results is publicly available at \texttt{github.com/skakouros/bert-prosody}. In the next we describe the BERT model architecture, related work, experimental methodology and results.

\section{BERT}
\vspace{-2mm}
BERT \cite{devlin2018bert} is a language model based on the Transformer architecture \cite{vaswani2017attention} that enables the bidirectional pre-training of representations by jointly conditioning on the left and right context in all layers. This allows the model to learn the entire surrounding context of a word which also means that the same word in different context will have a distinct representation. In contrast, earlier approaches looked at text sequences from left to right or by combining left to right and right to left training. The BERT representations are optimized based on two training objectives: (i) predicting randomly masked words in the input, and (ii) predicting whether the next sentence is the subsequent sentence in the input.

Our experiments are based on the \texttt{bert-base-uncased} variant of BERT. The model consists of 12 layers, each with an embedding size of 768, and 12 attention heads.

\section{Related Work}
\vspace{-2mm}
Probing network layers to investigate the structural knowledge of the language that a model has captured is an active research area that falls under the topic of neural network interpretability. In recent years there has been an increasing number of studies examining the representations that language models learn. Some works use probing tasks to unveil the linguistic features encoded in neural models \cite{jawahar2019does, tenney2019you}, others use attribution methods such as Integrated Gradients \cite{ramnath2020towards,nayak2021using} and some analyze Transformers' attention heads for evidence of linguistic and syntactic phenomena \cite{clark2019does}. In this work, we identify each layer's contribution to a specific prosodic task by attaching one trainable weight on each layer of BERT and training a light-weight classification head on top of a frozen pre-trained BERT. This enables us to observe the weight allocations across the layers and compare them with existing findings.

\section{Layer-wise Analysis}
\vspace{-2mm}
\subsection{Layer weights}
\begin{figure}[!t]
\begin{center}
\includegraphics[width=\linewidth]{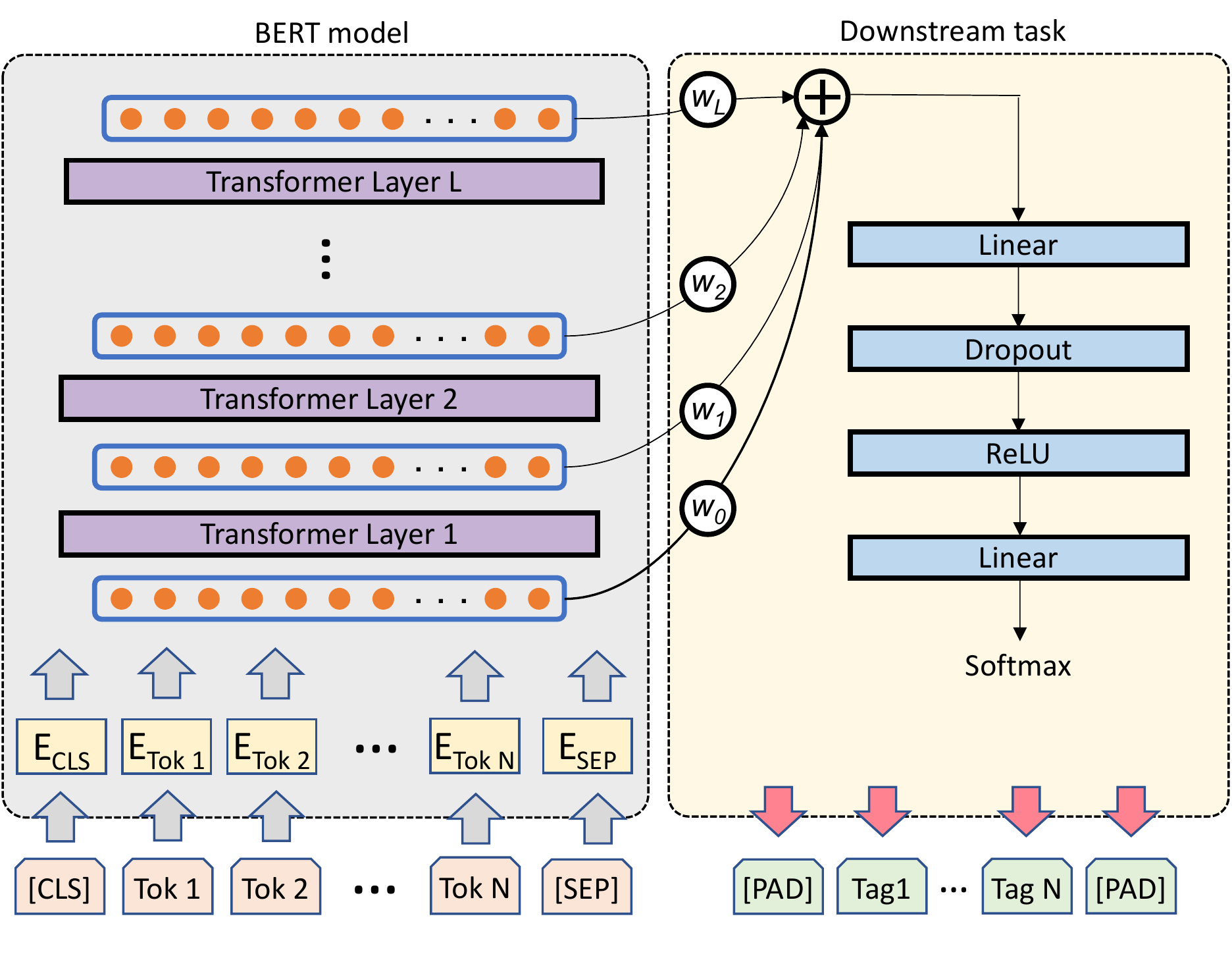}
\caption{Overview of the experimental setup.}
\vspace{-3mm}
\label{fig:setup_overview}
\end{center}
\end{figure}

\vspace{-2mm}
We obtain the contribution of BERT layers to the prediction task by introducing learnable scalar weights attached to each transformer layer of the model. An overview of the experimental setup can be seen in Fig.~\ref{fig:setup_overview}. We take representations from all transformer layers in the model and collapse them to one via a weighted average. There is one weight for each layer (a total of $L+1$) and all weights are trained jointly with the classification network. The weights are implemented with a learnable vector of size $L+1$, followed by the Softmax function.

\subsection{Layer embeddings}
\vspace{-2mm}
To further probe the contribution of individual BERT layers, we extract the embeddings from each layer separately and use them to train a classification head consisting of a single dense layer followed by the Softmax function. For each embedding layer, we then obtain the classification accuracy for the task.

\section{Experiments}
\vspace{-2mm}
In our experiments we use three datasets: two consisting of read speech and one of spontaneous dialogue speech. These are presented next followed by a description of the experimental setup.

\subsection{Data}
\vspace{-2mm}
\subsubsection{BURNC}
\vspace{-2mm}
The Boston University Radio News Corpus (BURNC) is a corpus of professionally read news data in American English \cite{ostendorf1995boston}. The corpus consists of speech from seven speakers (three female). The corpus also contains phonetic alignments, orthographic transcriptions, part-of-speech tags, and prosodic labels. In this work we use the text prompts and prosodic labels from the manually labelled part of the corpus (six speakers; approximately 3 h of data). The prosodic labeling system in BURNC is based on the Tones and Breaks Indices (ToBI) labeling convention and includes prosodic phrasing, phrasal prominence, and boundary tones. To obtain a single prominence label, all ToBI pitch accent types (e.g., H*,L*,L*+H) were marked as prominent while the rest as non-prominent.

\subsubsection{NXT Switchboard}
\vspace{-2mm}
The NXT Switchboard corpus is a dataset that includes the original Switchboard corpus annotations \cite{godfrey1992switchboard} into one coherent integrated format (NITE XML; NXT) enriched with annotations of prosody and contrast as well as syllable and phone information \cite{calhoun2010nxt}. The prosody annotations are available for a subset of the data that includes 76 conversations labelled with the ToBI transcribing convention. Similar to BURNC dataset, we use a binary prominence distinction marking words with ToBI accents as prominent and the rest as non-prominent.

\begin{figure}[!t]
\begin{center}
\includegraphics[width=\linewidth]{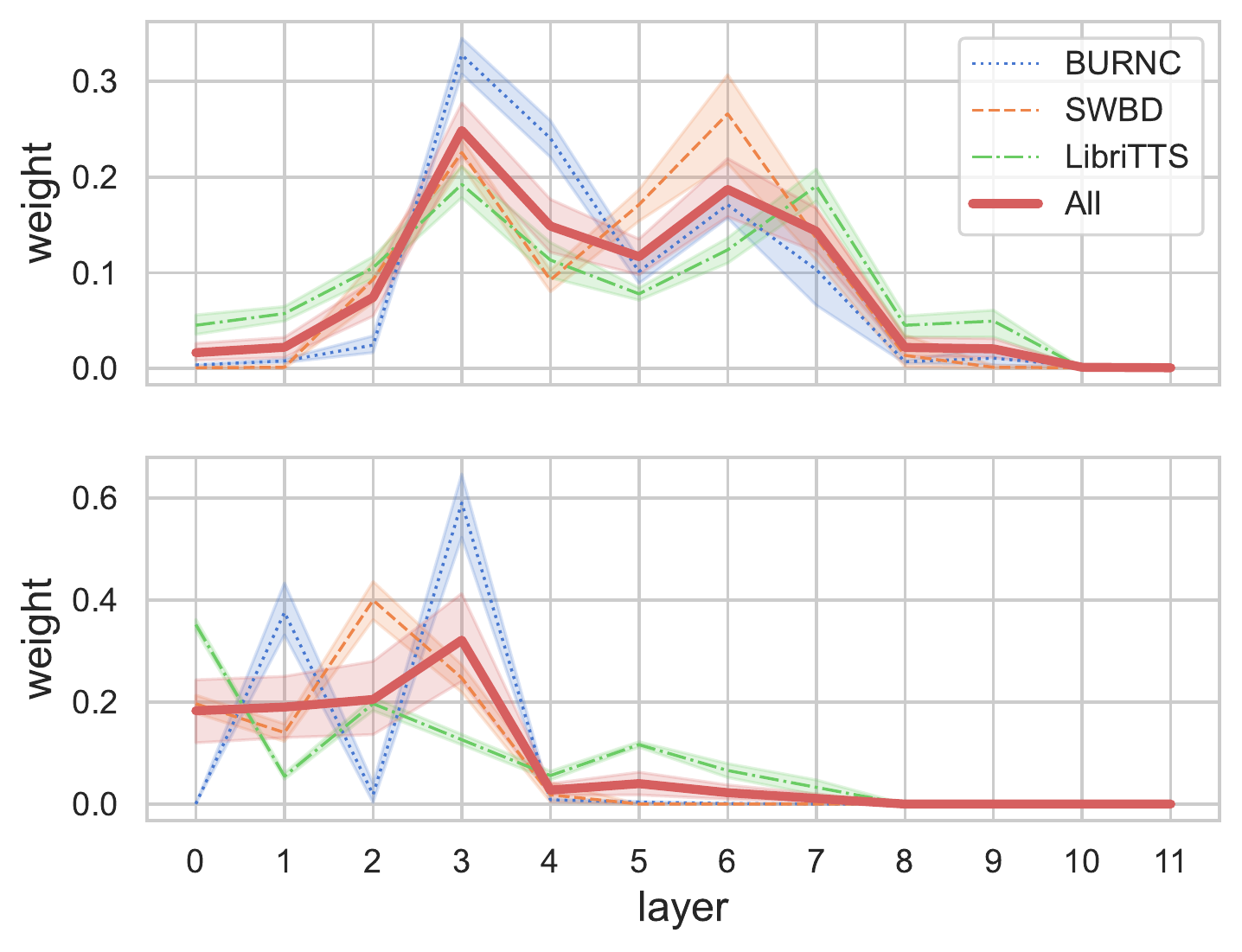}
\caption{BERT layer weights for prominence (top) and POS (bottom) prediction.}
\label{fig:results_weights}
\vspace{-3mm}
\end{center}
\end{figure}

\subsubsection{LibriTTS}
\vspace{-2mm}
The LibriTTS corpus \cite{zen2019libritts} is a processed (automatically aligned, segmented, and filtered) subset of the original audio and text data of the LibriSpeech corpus \cite{panayotov2015librispeech} that is based on English audiobooks of the LibriVox project. From the corpus we use the \textit{clean} subset that consists of 262.5 hours of read speech from 1230 speakers that were subsequently automatically labelled for prominence in the Helsinki Prosody Corpus (HPC) \cite{talman_etal2019prosody}. From the HPC data, we use the binary prominence tags.

\subsection{Experimental Setup}
\vspace{-2mm}
To obtain the layer weights, we train the network (only the classification head and layer weights) with frozen pre-trained BERT for prominence and POS prediction for each dataset separately. For the training we use a 80-15-5 split for train, validation, and test. We run the training for 20 epochs with a batch size of 4 and we repeat each experiment five times. The results are averaged across the independent runs for each task. The network is configured with a learning rate of $5e-5$ and a different learning rate $1e-2$ for the layers weights. This was done in order to allow the model to adjust the weights for the different layers more rapidly. We extract the weights from the model checkpoint with the best development accuracy and average them over the five runs. We repeat the same procedure and setup for POS prediction. 

In addition to the frozen pre-trained BERT we also fine-tune the entire model and report the results in Table~\ref{tab:promposresults} to compare the overall model performance with both fine-tuned and frozen BERT models. We use the same learning rates and epochs as the previous experiment with frozen BERT.

For POS prediction we used Spacy \cite{spacy3} to extract the part of speech categories. This resulted into 17 discrete classes. POS classes include, for example, adjectives, adpositions, adverbs (see \cite{spacy3} for a complete list of the coarse POS categories).

Finally, to balance the data and enable better comparison, each dataset is post-processed to include one full sentence per sample. As BURNC may include an entire paragraph and Switchboard several sentences per dialogue turn within a sample, we explicitly set sample size to be one sentence. Thus, a batch size of four includes four sentences.

\section{Results and Discussion}
\vspace{-2mm}
We report overall model performance for POS and prominence prediction in Table~\ref{tab:promposresults}, layer-specific performance in Table~\ref{tab:layerresults}, and illustrate how layer weights vary with respect to different tasks in Fig.~\ref{fig:results_weights}. We are interested in examining how different layers contribute to prominence and POS prediction with respect to findings on other tasks that have indicated different linguistic functions associated with different layers. Overall, for prominence we observe widespread distribution of the layer weights while POS appears more focused in the earlier BERT layers. These are presented in more detail in the next sections.

\subsection{Prosodic Prominence}
\vspace{-2mm}

\begin{table}[tbp]
\scriptsize 
\centering
\caption{Layer-wise accuracy for the test set runs for prominence and POS prediction with frozen BERT. Numbers in bold denote the two top results in each task.}
\vspace{-2mm}
\label{tab:layerresults}
\begin{tabular}{ *{7}{c}} 
\hline
\multirow{2}{2em}{\textit{Layer}} & \multicolumn{2}{c}{\textbf{BURNC}} & \multicolumn{2}{c}{\textbf{SWBD}} & \multicolumn{2}{c}{\textbf{LibriTTS}} \\
  & \textit{Prom} & \textit{POS} & \textit{Prom} & \textit{POS} & \textit{Prom} & \textit{POS} \\
\hline
 0 & 81.51 & 93.84 & 75.32 & \textbf{94.62} & 80.18 & 89.50\\
 1 & 83.52 & \textbf{93.90} & \textbf{76.14} & \textbf{94.67} & \textbf{80.32} & \textbf{90.08}\\
 2 & 85.20 & \textbf{94.33} & \textbf{76.01} & 94.02 & 80.29 & \textbf{89.53}\\
 3 & 85.20 & 93.84 & 75.47 & 93.50 & \textbf{81.08} & 89.00\\
 4 & \textbf{85.26} & 93.47 & 75.25 & 92.52 & 80.28 & 88.40\\
 5 & \textbf{85.26} & 93.17 & 75.82 & 91.87 & 80.00 & 87.29\\
 6 & 84.13 & 92.98 & 75.71 & 90.95 & 79.71 & 86.12\\
 7 & 84.46 & 91.76 & 75.82 & 90.11 & 79.40 & 84.86\\
 8 & 83.79 & 91.03 & 75.45 & 88.37 & 79.05 & 83.48\\
 9 & 82.92 & 89.20 & 74.71 & 87.20 & 78.63 & 82.36\\
 10 & 82.59 & 88.47 & 74.29 & 86.64 & 78.32 & 81.14\\
 11 & 83.26 & 87.19 & 73.08 & 82.09 & 77.27 & 76.67\\
\hline
\end{tabular}
\vspace{+1mm}
\end{table}

For prosodic prominence, the three datasets tested have shown differences in their overall performance. For example, for frozen BERT, prominence prediction accuracy was 87.14\% for BURNC, 78.10\% for Switchboard, and 82.67\% for LibriTTS. These differences are likely due to the different speaking styles involved in the data, with dialogue speech having the lowest performance. Another interesting observation in the results is that the model performance is degraded for the fine-tuned models. It seems that when fine-tuning the model, generalizability decreases due to overfitting on the idiosyncrasies of the training data. With the frozen model, performance is high and on a par with results reported in the literature. For instance, for LibriTTS \cite{talman_etal2019prosody} report 83.20\% and in our setup we obtained 82.67\%.

When it comes to layer weights, prosodic prominence has a widespread pattern of weights across BERT layers. Weights span most layers and are primarily focused within layers 2-8 with a peak appearing at layer 3. Interestingly, the pattern of weights appears to be quite similar across the three datasets tested. The spread of the weights suggests that different types of linguistic information are used in the prediction of prominent tokens. One interpretation of the results is that the model relies greatly on surface linguistic features such as POS but also involves syntactic and semantic information with increasing layers (see also \cite{jawahar2019does}). For BURNC, we also observe that the third layer has a high weight for both prominence and POS prediction. It is possible that prominence in BURNC (professionaly read speech) correlated more with POS than audiobooks (LibriTTS) and dialogue speech (SWBD). Another possibility for these differences could be also attributed to the different prominence coding schemes in the three datasets.

\subsection{POS}
\vspace{-2mm}
Part-of-speech information seems to be encoded in the early BERT layers with model accuracy being high for both frozen and fine-tuned runs of the experiments and across all datasets tested. Fine-tuning the model leads to improved performance, where for BURNC we get an increase in accuracy from 95.97\% with the frozen model to 97.56\% with fine-tuning. Switcboard and LibriTTS perform similarly with an increase in performance when the model is fine-tuned.

Layer weights for POS demonstrate a very different pattern when compared to prominence. POS information is found mainly in the lower layers of BERT with weights across the datasets varying but being focused in the early layers of the model. Most of the information seems to come from layers 0-4 which have been shown to encode surface features \cite{jawahar2019does}. This finding is also in agreement with other work that shows maximum POS tagging performance in the lower layers of models with accuracies ranging from 97.2\% to 97.4\% \cite{peters2018dissecting, peters-etal-2018-deep}.

\begin{table}[tbp]
\small 
\centering
\caption{Accuracy for the test set runs for prominence and POS prediction with frozen and fine-tuned (ft) BERT.}
\vspace{-2mm}
\label{tab:promposresults}
\begin{tabular}{lccc} 
\hline
\textit{Prominence} & \textbf{BURNC} & \textbf{SWBD} & \textbf{LibriTTS} \\
\hline
 frozen & 87.14 & 78.10 & 82.67 \\
 ft & 85.53 & 75.95 & 80.32 \\
\hline
\textit{POS} & \textbf{} & \textbf{} & \textbf{} \\
\hline
 frozen & 95.97 & 97.94 & 98.49 \\
 ft & 97.56 & 98.54 & 98.96 \\
 \hline
\end{tabular}
\vspace{+1mm}
\end{table}

\section{Conclusions}
\vspace{-2mm}
In this work, we performed a series of experiments on prosodic prominence to investigate whether prosody is part of the structural information of the language that BERT learns. Our results show that BERT captures information about prosodic prominence through a widespread allocation of weights across its layers reaching high performance. The weight allocations suggest that BERT relies on a variety of linguistic information for its predictions including surface linguistic features such as POS but also involving syntactic and semantic information. In future work, we will explore the same tasks with an extended set of datasets and methodological approaches. In addition to layer weights, we want to include layer integrated gradients as an attribution method in the experiments. We also aim to examine differences between the styles in the datasets, that is, read versus dialogue speech.

\section{Acknowledgements}
\vspace{-2mm}
This work was supported by the Academy of Finland project no. 340125 "Computational Modeling of Prosody in Speech" and Horizon 2020 Marie Sklodowska-Curie grant agreement No 859588. The authors wish to acknowledge CSC - IT Center for Science, Finland, for providing the computational resources.

\bibliographystyle{IEEEtran}
\bibliography{icphs2023}

\end{document}